\documentclass[a4paper, 10pt, conference]{ieeeconf}  
\usepackage{FG2018}

\FGfinalcopy % *** Uncomment this line for the final submission

\IEEEoverridecommandlockouts 

\usepackage{graphicx} % for pdf, bitmapped graphics files
\usepackage{amsmath} % assumes amsmath package installed
\usepackage{amssymb}  % assumes amsmath package installed
\usepackage{url}
\usepackage{balance}

\title{\LARGE \bf
A study on the use of Boundary Equilibrium GAN for Approximate Frontalization of Unconstrained Faces to aid in Surveillance
}

\author{\parbox{16cm}{\centering
    {\large Wazeer Zulfikar, Sebastin Santy, Sahith Dambekodi and Tirtharaj Dash}\\
    {\normalsize BITS Pilani - KK Birla Goa Campus, Goa, India}\\
    {\normalsize \{f20150003, f20150357, f20150192, tirtharaj\}@goa.bits-pilani.ac.in}
    }
}

\begin{document}

\maketitle

%%%%%%%%%%%%%%%%%%%%%%%%%%%%%%%%%%%%%%%%%%%%%%%%%%%%%%%%%%%%%%%%%%%%%%%%%%%%%%%%
\begin{abstract}
Face frontalization is the process of synthesizing frontal facing views of faces given its angled poses. We implement a generative adversarial network (GAN) with spherical linear interpolation (Slerp) for frontalization of unconstrained facial images. Our special focus is intended towards the generation of approximate frontal faces of the side posed images captured from surveillance cameras. Specifically, the present work is a comprehensive study on the implementation of an auto-encoder based Boundary Equilibrium GAN (BEGAN) to generate frontal faces using an interpolation of a side view face and its mirrored view. To increase the quality of the interpolated output we implement a BEGAN with Slerp. This approach could produce a promising output along with a faster and more stable training for the model. The BEGAN model additionally has a balanced generator-discriminator combination, which prevents mode collapse along with a global convergence measure. It is expected that such an approximate face generation model would be able to replace face composites used in surveillance and crime detection.
\end{abstract}

%%%%%%%%%%%%%%%%%%%%%%%%%%%%%%%%%%%%%%%%%%%%%%%%%%%%%%%%%%%%%%%%%%%%%%%%%%%%%%%%
\section{Introduction}
\label{sec:intro}
For various security and surveillance applications, human faces have been considered as one of the most universally used biometry identification tool along with other biometry features such as fingerprints, signature and so forth~\cite{chopra2005learning}. However, there exists an underlying assumption that faces of two different subjects can not be same, which may not hold for faces of twin subjects for many different biological reasons. The main interest of the present work is not handling such biological difficulties rather a work on the application of contemporary machine learning (ML) tools towards face generation from incomplete face images~\cite{park2010face,sinha2006face}. Accurate identification of faces often require the whole frontal view of the subject's face, which is not available in certain circumstances such as surveillance camera footage or when the subject is dynamic that changes the orientation of the face. To aid with the facial identification process in these circumstances, frontalization of human faces has widely been used on the angled views of the faces~\cite{mallikarjun2015efficient,sagonas2015face}. Face Frontalization is the process of synthesizing frontal facing view from the angled pose of the human face in which one side of the face is not clearly visible or not available at all. In the fast growing area of automatic face and gesture recognition research, face frontalization is one of the most important and discussed problems of fundamental interest in both human and machine facial processing and recognition.

Frontal face images can be automatically generated using one of the popular deep learning approaches called Generative Adversarial Networks (GAN)~\cite{goodfellow2014generative}. GAN are a class of deep learners that are structured around two functions: the generator $\mathcal{G}(\mathbf{z})$, which maps a sample $\mathbf{z}$ from a Gaussian distribution (also called noise) to the data distribution, and the discriminator, denoted as $\mathcal{D}(\mathbf{x})$, which determines if a sample $\mathbf{x}$ belongs to the data distribution. The generator and discriminator are typically learned jointly by alternating the training of $\mathcal{D}$ and $\mathcal{G}$, The two learner models essentially contest with each other in a notional framework of zero-sum game. GAN models have shown remarkable prediction accuracy, generator-discriminator power balance, faster convergence, and higher visual quality, they have been employed in many different face and gesture recognition applications~\cite{ledig2016photo}. Most recently, a modified model based on auto-encoder based Boundary Equilibrium GAN (BEGAN) was proposed for a similar task of automatically face synthesis~\cite{berthelot2017began}.

Our present paper closely follows the Berthelot et al.'s BEGAN work for a comprehensive study on the automatic face frontalization for human faces. Specifically, we use single side-posed face image to generate frontal face using BEGAN with a different kind of interpolation setting. This study clearly suggests that a large dataset of frontal face might not be needed in application to surveillance jobs and would definitely aid in saving time and adding more security in various ways. Throughout the work and of biological interest, we assume that faces ought to be symmetrical with expected negligible deviations.

The rest of the paper has been organized as follows: section \ref{sec:intro} discussed the introduction and the problem statement, section \ref{sec:metho} depicts the methodology used in this work. Section \ref{sec:experiment} explains the experimental setup and the evaluated results. Some of the related works are presented in section \ref{sec:relwork}. The paper is concluded in section \ref{sec:concl}.

\section{Methodology}
\label{sec:metho}
In this work, the problem has been approached by exploiting the interpolatory feature of GAN~\cite{chen2016infogan}. As mentioned in the introduction section, the methodology would closely follow BEGAN experiment\cite{berthelot2017began} with a different kind of interpolation strategy. Boundary Equilibrium Generative Adversarial Networks (BEGAN) use an auto-encoder as the discriminator $\mathcal{D}(\cdot)$. The primary motive for this is to change the loss function. The reconstruction loss of the auto-encoder is used instead of the real loss of the images. The real loss used for training is then derived from the Wasserstein distance~\cite{ruschendorf1985wasserstein} between the reconstruction losses. A hyperparameter, $\gamma$ controls the modal collapse by balancing losses used for training the discriminator and the generator. BEGAN also offers a quantitative convergence measure to track the quality of timely convergence of the model.

For synthesizing the specific frontal view of the subject face, the side view and its mirrored image are interpolated. First the images are encoded using the discriminator $\mathcal{D}(\cdot)$, which is an auto-encoder to generate the individual embeddings ($\mathbf{z}_0$ and $\mathbf{z}_1$). In this work, these embeddings $\mathbf{z}_0$ and $\mathbf{z}_1$ are then interpolated using a technique called Spherical Linear Interpolation, known as Slerp---a kind of quaternion interpolation~\cite{kremer2008quaternions}---that could able to show promising performance on the discussed problem of face frotnalization. The interpolated embeddings serve as input to the generator $\mathcal{G}(\cdot)$ to produce the frontal view of the face.

In the present implementation, the two side views of the face are taken as the endpoints of the arc. The points in the arc correspond to the intermediate images. Let $\mathbf{z}_0$ and $\mathbf{z}_1$ be the first and last points of the arc, and let $t$ be the parameter, $t \in [0,1]$. Compute $\Omega$ as the angle subtended by the arc so that $ \cos\Omega = \mathbf{z}_0\cdot \mathbf{z}_1$. Mathematically, Slerp could be written as a formula:
\begin{equation}
\mathsf{Slerp}(\mathbf{z}_0,\mathbf{z}_1; t) = \frac{\sin {[(1-t)\Omega}]}{\sin \Omega} \mathbf{z}_0 + \frac{\sin [t\Omega]}{\sin \Omega} \mathbf{z}_1
\label{eq:slerp}
\end{equation}

As $ \Omega \to 0$ , the formula in Equation \eqref{eq:slerp} reduces to the linear interpolation formula.
\begin{equation}
\mathsf{Lerp}(\mathbf{z}_0,\mathbf{z}_1; t) = (1-t) \mathbf{z}_0 + t \mathbf{z}_1
\label{eq:lerp}
\end{equation}

By varying the value of $t$, we can effectively plot a point between the two points on the arc. This formula is particularly useful for 3-dimensional rotations like that of the faces being generated. Given two faces $\mathbf{z}_0$ and $\mathbf{z}_0$, the rotation of one face to match the orientation of the other is done using the Slerp formula in Equation \eqref{eq:slerp}. The parameter $t \in [0,1]$ is used to rotate in steps so that accuracy could improve. So, at each change in the value of $t$, the face is rotated slightly towards the desired orientation effectively by plotting a path for interpolation between the two faces.

\begin{figure}[h]
	\centering
	\includegraphics[scale=0.5]{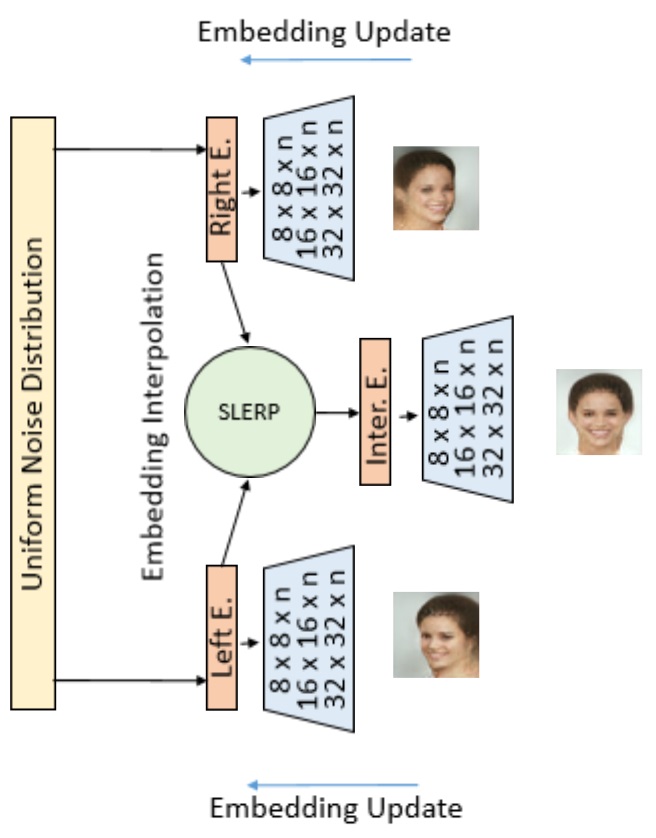}
	\caption{Demonstration of embedding interpolation (The Right E. (Right Embeddings) and the Left E. (Left Embeddings) which are sampled from the uniform distribution are first trained using feedback from Real Images. Inter. E. (Interpolated Embeddings) are formed using Slerp interpolation. The Inter. E. is passed through the trained generator to produce output. The trained generator consists of 3 de-convolutional layers (input size of 8, 16 and 32) with an up-sampling of 2 each.)
	} \label{fig:embnet}
\end{figure}

We demonstrate the concept of embedding interpolation in the framework of face frontalization in figure \ref{fig:embnet}. 
To achieve frontalization we start with the faces slightly panned by 45-50 degrees towards the side. Interpolation by the GAN will rotate the angle of the face to 0 degrees. Firstly, a random uniform noise distribution is sampled twice to obtain rudimentary embeddings for the left and right views. The embeddings are passed through the generator to synthesize an image. The gradients of the construction loss is used to update the embeddings of the side views. The updated embeddings are interpolated using spherical linear interpolation to get the optimal embeddings for generating the frontal view. These optimal embeddings are passed through the generator to synthesize the final frontal view of the face.

\section{Experiment}
\label{sec:experiment}
\subsection{Setup}
We trained our present model of BEGAN with Slerp using the Adam Optimizer \cite{kingma2014adam} with a small learning rate of 0.001. Such a small learning rate is preferred so as to avoid any kind of modal collapse or memory based solutions~\cite{berthelot2017began}. The BEGAN with Slerp model was trained with CelebA dataset that contains $64\times 64$ images\cite{liu2015faceattributes} in order to minimize the training duration. The generator $\mathcal{G}(\cdot)$ was a deconvolution network, while the discriminator $\mathcal{D}(\cdot)$ was an auto-encoder comprised of a convolution decoder and a deconvolution encoder. Training time was about 36 hours on one Tesla K80 GPU\footnote{\url{https://www.floydhub.com/}} for 112k Epoch timeout.

The interpolation setup is as follows. A batch size of 64 was used where the first 32 batches contained the original faces with angles in the range of $20^\circ$ to $60^\circ$. The next 32 batches comprised of the corresponding left-right flipped mirror views. The embeddings were trained to match with the given face image. The parameter $t$, which controls the point clusters on the arc is then varied from 0 to 1 (where 0 and 1 indicate the extreme ends of the interpolation) in uniform intervals of 0.1. However, this can be varied by following the equation \eqref{eq:nt} and \eqref{eq:n} where, $\left \lceil t \right \rceil$ is the upper bound of $t$, $\left \lfloor t \right \rfloor$ is the lower bound of $t$, $\Delta$ is the interval size, and $n$ is the total number of intermediate angle of each image. For every value of $t$, the embedding vector was taken to generate the corresponding intermediate image using the generator. This will lead to 8 uniform angle deviations between any image and its mirrored view. So, a total of 10 images are produced as the output. The median two images are the frontal faces.
\begin{equation}
n_t = \frac{\left \lceil t \right \rceil - \left \lfloor t \right \rfloor}{\Delta}
\label{eq:nt}
\end{equation}
\begin{equation}
{n} = \frac {angle_{left} - angle_{right}} {n_t}
\label{eq:n}
\end{equation}

\subsection{Evaluation}
\begin{figure*}[h]
	\centering
	\includegraphics[scale=0.65]{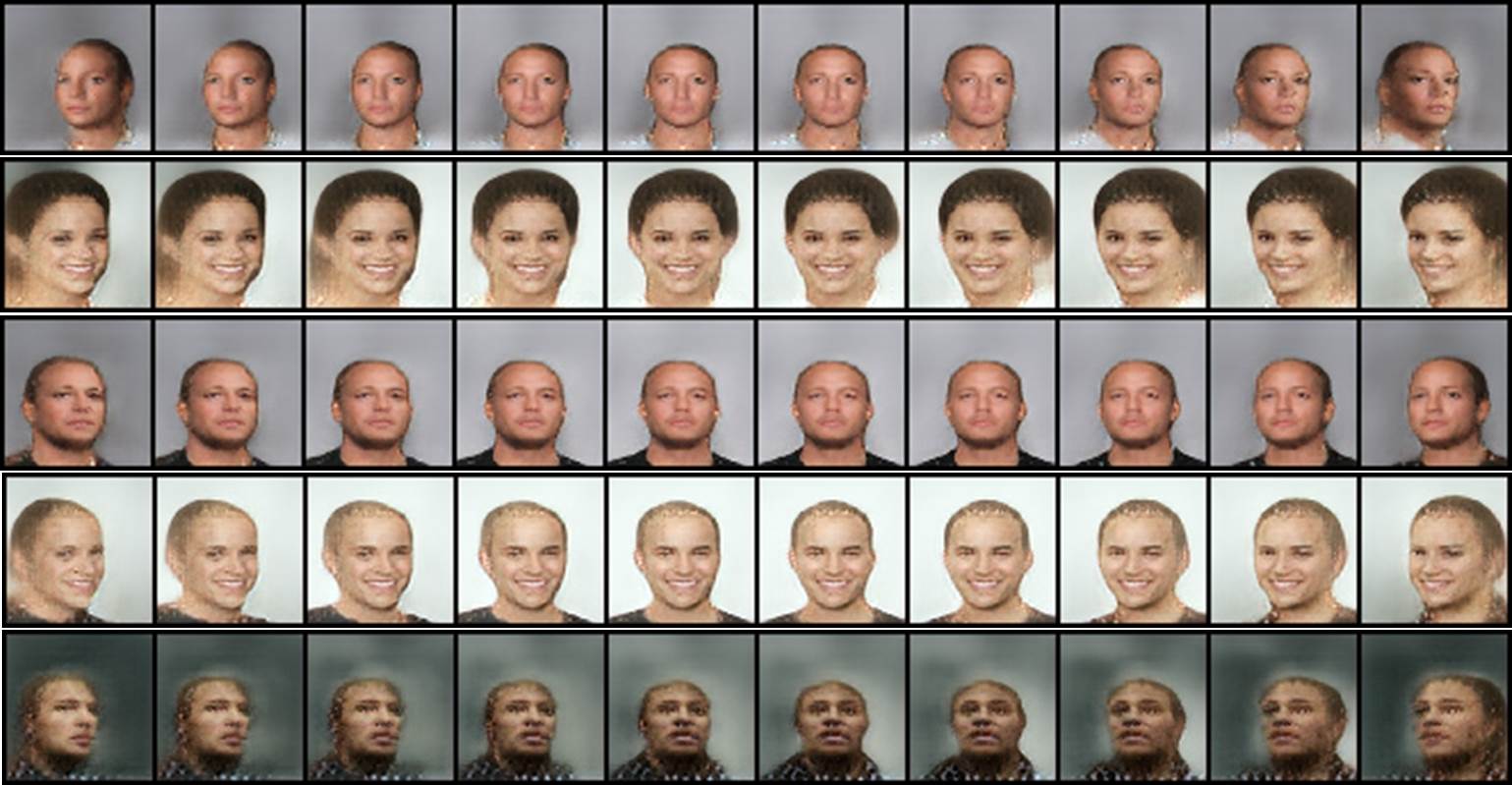}
	\caption{Face frontalization using BEGAN with Slerp on images of five different subjects (It should be noted that the results shown here are not of great quality due to the limited availability of computational resources. Howver, the median images show the generated front faces from the angled pose of the face.)} \label{fig:results}
\end{figure*}

We first start with the the left-most image. We mirror this image and then start interpolation. In some cases there is a slight difference between the original image and its mirrored image, which are present on the extreme ends. This difference occurs because of the way the generation of the faces occurs. First, the mirroring occurs and the faces are then fed into the network. The network then generates the faces again with the loss. Since the generation of faces has some randomization from the  uniform random distribution there is a chance of slight deviations between the images. However, this doesn't make a large difference on the final result except for a slight difference in certain facial features which are different between the two images. The results obtained from our experiment are shown in the figure \ref{fig:results}.

\section{Related Works}
\label{sec:relwork}
Some of the previous works on face frontalization have mapped the original side posed image to 2-dimensional projection using mathematical approximations~\cite{mallikarjun2015efficient,hassner2015effective}. These methods use 3D surface as an approximation to the shape of all input faces, and project them point-wise. A recent work on Face Frontalization using GAN (FFGAN) takes the help of 3D morphable model coefficients to aid with faster training and better convergence~\cite{ferrari2016effective}.

\section{Concluding Remarks}
\label{sec:concl}
This paper presents a study on the implementation of Boundary Equilibrium GAN with Spherical Linear Interpolation (Slerp) for the problem of face frontalization. This work could aid to the face synthesis problem in surveillance applications in which the images from the footage are mostly noisy and faces are angled. The present work demonstrated promising results on a limited resource system. However, there are few future extensions that could be done based on the present study:
\begin{itemize}
	\item A primal assumption made in our methods is that the original face of the person is symmetric in nature. Since there is no way to definitively know whether the face is symmetric using just one image, prior domain knowledge would aid in generating accurate frontal views for asymmetrical faces as well.
	\item As also mentioned in previous articles on GAN, the discriminator model could be designed as a variational auto-encoder that could probably improve the performance.
\end{itemize}

%%%%%%%%%%%%%%%%%%%%%%%%%%%%%%%%%%%%%%%%%%%%%%%%%%%%%%%%%%%%%%%%%%%%%%%%%%%%%%%%
\section{Acknowledgment}

Acknowledge shall be added after the FG 2018 peer-review and acceptance.

%%%%%%%%%%%%%%%%%%%%%%%%%%%%%%%%%%%%%%%%%%%%%%%%%%%%%%%%%%%%%%%%%%%%%%%%%%%%%%%%

\bibliographystyle{unsrt}
\bibliography{refs}

\balance

\end{document}